\documentclass[runningheads]{llncs}

\usepackage{eccv}

\usepackage{eccvabbrv}

\usepackage{graphicx}
\usepackage{booktabs}

\usepackage[accsupp]{axessibility}  

\usepackage[pagebackref,breaklinks,colorlinks,citecolor=eccvblue]{hyperref}

\usepackage{orcidlink}

\usepackage[export]{adjustbox} 
\renewcommand{\paragraph}[1]{\vspace{1.25mm}\noindent\textbf{#1}}

\usepackage{tabulary,array}
\newcolumntype{x}[1]{>{\centering\arraybackslash}p{#1pt}}

\usepackage{sidecap}
\usepackage{multirow}

\begin{document}

\title{UniProcessor: A Text-induced Unified Low-level Image Processor} 

\titlerunning{UniProcessor}

\author{Huiyu Duan\inst{1,2}\orcidlink{0000-0002-6519-4067} \and
Xiongkuo Min\inst{1,\star}\orcidlink{0000-0001-5693-0416} \and
Sijing Wu\inst{1} \and \\
Wei Shen\inst{2,\star} \and
Guangtao Zhai\inst{1,2,}\thanks{Corresponding authors.\\This work was supported in part by the National Key R\&D Program of China 2021YFE0206700; in part by the NSFC 62225112, 62322604, 62176159, 62271312; in part by the Shanghai Pujiang Program 22PJ1407400.}\orcidlink{0000-0001-8165-9322}
}

\authorrunning{Huiyu Duan et al.}

\institute{Institute of Image Communication and Network Engineering, \\Shanghai Jiao Tong University \and
MoE Key Lab of Artificial Intelligence, AI Institute, Shanghai Jiao Tong University
\email{\{huiyuduan,minxiongkuo,wusijing,wei.shen,zhaiguangtao\}@sjtu.edu.cn}
}

\maketitle


\begin{abstract}
Image processing, including image restoration, image enhancement, \textit{etc.}, involves generating a high-quality clean image from a degraded input.
Deep learning-based methods have shown superior performance for various image processing tasks in terms of single-task conditions.
However, they require to train separate models for different degradations and levels, which limits the generalization abilities of these models and restricts their applications in real-world.
In this paper, we propose a text-induced \underline{Uni}fied image \underline{Processor} for low-level vision tasks, termed \textbf{UniProcessor}, which can effectively process various degradation types and levels, and support multimodal control.
Specifically, our UniProcessor encodes degradation-specific information with the subject prompt and process degradations with the manipulation prompt.
These context control features are injected into the UniProcessor backbone via cross-attention to control the processing procedure.
For automatic subject-prompt generation, we further build a vision-language model for general-purpose low-level degradation perception via instruction tuning techniques.
Our UniProcessor covers 30 degradation types, and extensive experiments demonstrate that our UniProcessor can well process these degradations without additional training or tuning and outperforms other competing methods.
Moreover, with the help of degradation-aware context control, our UniProcessor first shows the ability to individually handle a single distortion in an image with multiple degradations.
Code is available at: \url{https://github.com/IntMeGroup/UniProcessor}.
\end{abstract}

\begin{figure*}[t]\centering
\vspace{-0.8em}
\includegraphics[width=0.98\linewidth]{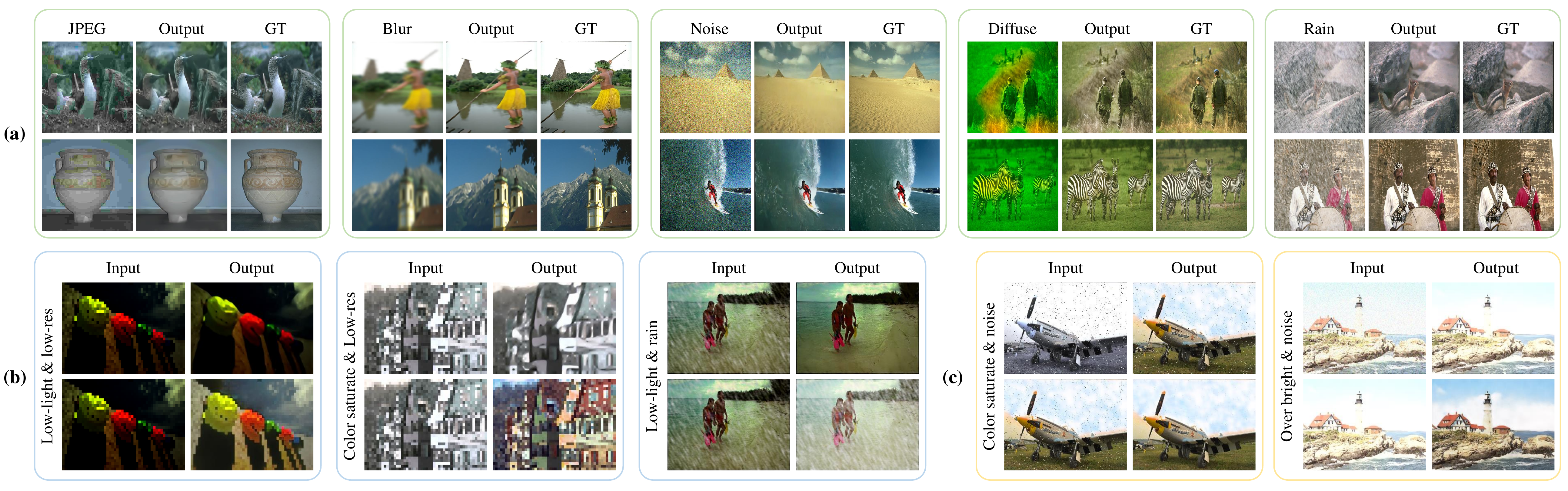}
\vspace{-0.6em}
\caption{
\textbf{UniProcessor} is capable of processing various degradations in one model with text control. (a) For single degradation, UniProcessor can well restore images. (b) For an image with multiple degradations, our UniProcessor can process individual distortion with text control, which demonstrates the superior distortion perception and disentangling abilities. (c) For images with multiple degradations, Uniprocessor can process each degradation step by step to restore or enhance the images.
}
\label{fig:1_frontpage}
\vspace{-1em}
\end{figure*}

\section{Introduction}
\label{sec:intro}

During image acquisition, storage, transmission, and rendering, degradations (such as noise, blur, rain, compression, \textit{etc.}) are often introduced, which significantly influence the quality of an image \cite{duan2023masked}.
Image processing, including image restoration, image enhancement, \textit{etc.}, aims at improving the quality of a degraded image and generating a high-quality clean output.
Due to the ill-posed nature, this problem is highly challenging and generally requires strong image priors for effective processing \cite{duan2023masked,zamir2022restormer}.
With the availability of large-scale training datasets, deep learning-based image processing methods have been widely developed to tackle various low-level vision tasks, such as denoising \cite{DnCNN,FFDNetPlus}, deblurring \cite{xu2013unnatural,deblurgan}, deraining \cite{zhang2019image,yang2017deep}, enhancement \cite{wei2018lol,wang2018gladnet}, \textit{etc.}, owing to its strong ability to learn generalizable image priors \cite{ulyanov2018dip,chen2021IPT,zamir2022restormer,duan2022develop}.

Many deep neural networks (DNNs) have been proposed to handle single low-level vision tasks \cite{ren2021adaptive,nah2021clean,ren2020single}.
These methods mainly incorporate task-specific features into the network to process a single problem, such as denoising \cite{zhang2017learning,ren2021adaptive}, deblurring \cite{zhang2020deblurring,nah2021clean}, \textit{etc.}, which lack the generalization ability to be used on other degradation processing tasks.
Some DNN-based methods have focused on designing a robust network architecture to tackle various distortions using one model \cite{duan2023masked,zamir2022restormer,tu2022maxim,wang2021uformer,chen2021IPT}.
Although using one network model, they need to train separate copies with different weights for solving different degradation types or degradation levels.
This limits the application of these models in practical scenarios due to the tedious process and complex deployment, and they need to select an appropriate pre-trained weight during the inference process, which requires prior knowledge and additional manipulations.

Recently, several methods towards handling multiple weather degradations using one model with one weight have been proposed \cite{liu2022tape,valanarasu2022transweather,li2020all,li2022all,potlapalli2023promptir,conde2024high}.
Some of these methods train parallel encoders or decoders for each specific weather degradation, which is hard to scale to more distortion types \cite{liu2022tape,valanarasu2022transweather,li2020all}.
AirNet \cite{li2022all} proposes an all-in-one restoration model by utilizing an extra encoder and employing contrastive learning to differentiate various corruption types.
However, it can not well disentangle different degradation representations \cite{potlapalli2023promptir}.
PromptIR \cite{potlapalli2023promptir} proposes to use learnable prompts to disentangle different degradation representations.
However, the learned prompt collection is a complete black-box, which is hard to understand and control.
Since in some cases, we may just want to tackle one specific distortion but ignore other degradations, it is important to develop a model that can well disentangle distortion representations and handle each degradation independently.
Moreover, these aforementioned models can only handle several degradations, which still have limitations in a wide range of practical applications.

In this paper, we propose an all-in-one text-induced \underline{Uni}fied image \underline{Processor} (\textbf{UniProcessor}) to tackle various low-level vision image-processing tasks.
Our method supports multimodal control, which first encodes input image and subject prompt to obtain degradation-specific information, \textit{i.e.}, subject prompt embedding, then this embedding and the manipulation prompt are fed into the text encoder to obtain the context control embedding for flexible image manipulation.
By interacting context control embedding with the feature representations of the main processor network, we dynamically control and enhance the representations with the degradation-specific knowledge and manipulation-aware information.
As shown in Fig. \ref{fig:1_frontpage}, benefiting from the explicit text induction, our UniProcessor can not only well restore images with a single degradation, but also separately or gradually process individual distortions in an image with multiple degradations using text control.
In order to facilitate interaction and control, we further develop a vision-language model for general-purpose low-level vision degradation perception via instruction tuning techniques, which can be used to automatically generate the subject prompt for text control.
Our UniProcessor is trained on 30 degradation types with various levels to conform to a variety of applications.
The main highlights of this work include:

\begin{itemize}
    \item We present a text-induced framework UniProceesor for all-in-one blind low-level image processing tasks, which has flexible and convenient text control ability.
    \item We propose a multimodal control module to achieve flexible control ability, which encodes degradation-specific information from the input image and subject prompt, and combines the obtained subject prompt embedding with the manipulation prompt to get the context control embedding.
    \item An effective processor backbone is developed, which contains a context interaction module to interact with the obtained context control embedding. 
    The multimodal control module and the context interaction module are plug-in modules, which can be easily integrated into any existing image processing network.
    \item For automatic subject-prompt generation, a vision-language model for general-purpose low-level degradation queries is devised via instruction tuning techniques.
    \item Beyond the state-of-the-art performance on single distortion processing, our UniProcessor can process multi-degradations individually, which manifests the superior degradation disentangling ability. Moreover, to the best of our knowledge, this is the first work that attempts to solve so many degradation problems using one network.
\end{itemize}

\section{Related Work}
\vspace{-0.1em}
\subsection{Image Processing}
\vspace{-0.1em}
With the development of deep learning techniques and the establishment of various databases and benchmarks, many DNN networks have been developed to handle various image restoration and image enhancement tasks, such as denoising \cite{DnCNN,FFDNetPlus}, deblurring \cite{xu2013unnatural,deblurgan}, deraining \cite{zhang2019image,yang2017deep}, low-light enhancement \cite{wei2018lol,wang2018gladnet}, \textit{etc.}, and have achieved state-of-the-art performance.
Many previous works have focused on the network design, and have proposed numerous robust architectures.
Some early studies have adopted convolutional neural network (CNN) as the backbone \cite{DnCNN,zamir2020mirnet,Zamir_2021_CVPR_mprnet,zhang2018image,zhang2020rdn}, and have devised many general-purpose or task-specific modules for various tasks, such as residual and dense connection \cite{DnCNN,ledig2017photo,wang2018esrgan,zhang2020rdn}, channel attention \cite{niu2020single,zamir2020mirnet,Zamir_2021_CVPR_mprnet,duan2022develop}, spatial attention \cite{wang2018non,liu2018non,Zamir_2021_CVPR_mprnet,duan2022develop}, multi-scale or multi-stage networks \cite{dmphn2019,deblurganv2,mspfn2020,cho2021rethinking_mimo,Zamir_2021_CVPR_mprnet,chen2021hinet}, \textit{etc.}
Recently, with the success of using transformer architecture across various computer vision (CV) tasks, many transformer-based networks have been developed to solve image processing problems, such as IPT \cite{chen2021IPT}, SwinIR \cite{liang2021swinir}, Uformer \cite{wang2021uformer}, Restormer \cite{zamir2022restormer}, CSformer \cite{duan2023masked}, \textit{etc}.
However, the aforementioned models can only solve one image restoration problem with one weight value, which lacks the generalization ability to be applied to various scenarios.
Some works have proposed unified models to tackle the images corrupted due to multiple weather conditions, such as snow, rain, haze \cite{liu2022tape,valanarasu2022transweather,li2020all}.
However, they need parallel multiple encoders or decoders for different tasks, which is hard to extend to more degradation types due to the dramatically increasing computational overhead.
AirNet \cite{li2022all} and PromptIR \cite{potlapalli2023promptir} are two recent methods towards achieving all-in-one image restoration using contrastive learning or prompt learning.
They are still methods that simply learn the mapping from various degraded domains to one clean domain.
However, in many cases, we may want to control each degradation separately, which is beyond the capabilities of the above models.

\vspace{-0.8em}
\subsection{Vision-language Model}
In recent years, many large-scale vision-language models have been proposed and have greatly promoted the development of the CV field.
Some vision-language models have tried to build foundation models for vision-language feature alignment.
CLIP \cite{radford2021learning} is an influential method that aligns image features and text features using contrastive learning.
FLIP \cite{li2023scaling} presents a mask pre-training method for the scaling vision-language pre-training process.
Based on the multimodal feature alignment pre-training methods, BLIP \cite{li2022blip} proposes a pre-training method for vision-language understanding and visual question answering (VQA).
Benefiting from the rapid evolution of large language models (LLMs) \cite{touvron2023llama}, BLIP-2 \cite{li2023blip2} presents to use frozen image encoders and LLMs, and only train a lightweight querying transformer to save training costs.
InstructBLIP \cite{instructblip} attempts to train general-purpose vision-language models based on BLIP-2 \cite{li2023blip2} and LLMs \cite{touvron2023llama} with instruction tuning.
Based on these general-purpose large-scale multimodal pre-training techniques, many text-to-image generation and editing methods have also been proposed \cite{rombach2022high,kang2023scaling,ruiz2023dreambooth,li2023blipdiffusion}.

\begin{figure*}[t]\centering
\vspace{-0.8em}
\includegraphics[width=0.96\linewidth]{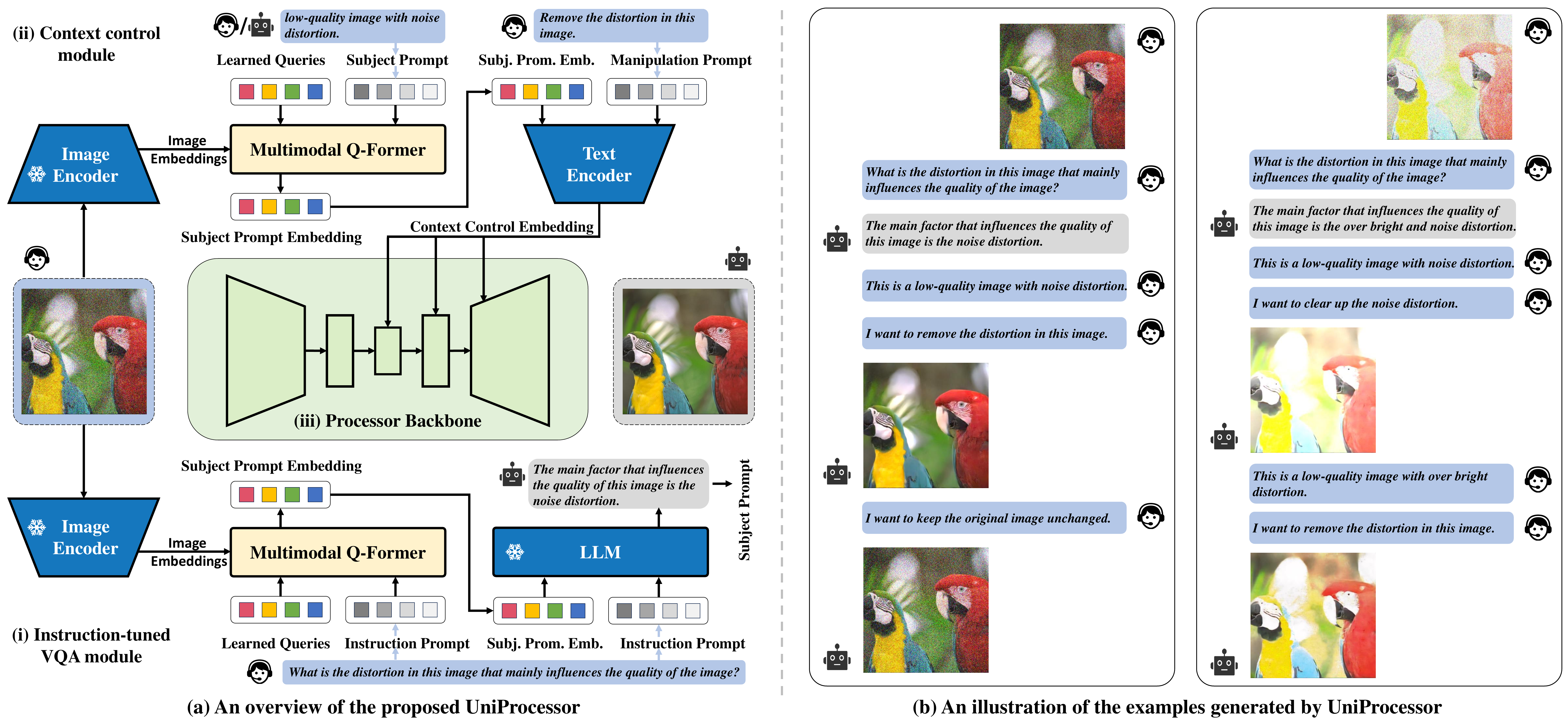}
\vspace{-0.6em}
\caption{An illustration of the overview and the examples of UniProcessor. 
(a) An overview of the proposed UniProcessor. 
(i) Our UniProcessor first learns low-level vision-language model via instruction tuning, which can adapt to various degradation-aware visual questions and generate the subject prompt. 
(ii) The subject prompt and the extracted input image embedding are encoded to obtain the subject prompt embedding, which is then combined with the manipulation prompt to obtain the context control embedding.
(iii) The guidance information is injected into the Processor backbone at multiple decoding stages.
(b) An illustration of the examples generated by Uniprocessor, which demonstrates the good control ability and degradation disentangling capability.
}
\label{fig:2_overview}
\vspace{-1em}
\end{figure*}

\section{Approach}

Given an input degraded image $\textbf{I} \in \mathbb{R}^{H \times W \times 3}$ with an unknown degradation $D$, we aim to learn a single model $M$ to process this image and obtain a high-quality output $\hat{I}$.
Fig. \ref{fig:2_overview} shows the overview and the examples of our UniProcessor.
Our UniProcessor mainly contains three parts, which include an instruction-tuned VQA module, a context control module, and a processor backbone.
Our UniProcessor supports human-in-loop manipulation with user-supplied text and automatic manipulation with the prompt generated by the instruction-tuned VQA module.
For automatic manipulation, the \textbf{overall pipeline} is given as follows.
For the degraded input $I$, UniProcessor first uses the instruction-tuned VQA module to automatically generate a subject prompt.
The subject prompt and the extracted input image embedding are then encoded using a multimodal Q-Former to obtain the subjective prompt embedding, which is then combined with the manipulation prompt using a text encoder to obtain the context control embedding.
Finally, the context control embedding is injected into the Processor backbone at three decoding stages to control the process procedure and generate the output restored image $\hat{I}$.

\vspace{-0.5em}
\subsection{Low-level Vision-language Instruction Tuning}
We develop an instruction-tuned VQA module for UniProcessor, which can adapt to various degradation-aware visual questions and generate the subject prompt.

\paragraph{Data preparation.}
In order to enable the low-level vision perception and instruction ability of our UniProcessor, we first establish a new VQA database for low-level vision.
We devise a distortion bank including 30 types of degradation (see supplementary material for more details).
Based on this distortion bank, we generate numerous degraded images with various corruption types and levels for over 70000 image patches.
Since the degradation types and levels for these images are known, we further generate various degradation-related or quality-related questions and answers for training the low-level vision VQA model.
To avoid the risk of model overfitting, we follow the InstructBlip \cite{instructblip} to craft 10 to 15 distinct instruction templates in natural language to articulate the task and the objective (see supplementary material for more details).

\paragraph{Instruction-aware visual feature extraction.}
UniProcessor first extracts instruction-aware visual features for feasible question answering.
As shown in Fig. \ref{fig:2_overview} (a)-(i), the instruction-tuned VQA module encodes the input degraded image with a well pre-trained  frozen CLIP image encoder \cite{radford2021learning} to obtain the image embedding.
Moreover, the instruction text is also encoded by a pre-trained text encoder to obtain the instruction prompt.
Then a multimodal Query Transformer, \textit{i.e.}, Q-Former, is utilized to extract instruction-aware visual features by jointly interacting image embedding, text prompt, and $K$ learnable query embeddings.
The output of the Q-Former consists of $K$ visual vectors, of which the number is the same as the learnable query embeddings.

\paragraph{Instruction-tuned low-level vision VQA.}
The above extracted instruction-aware visual features are then fed into a frozen LLM as soft prompt input to perform instruction-guided VQA.
The frozen LLM adopted in UniProcessor is LLaMA \cite{touvron2023llama}.
The connection between the Q-Former and LLM is a fully-connected layer, which adapts the output instruction-aware visual features of the Q-Former to the input dimension of the LLM.
The Q-Former is pre-trained with InstructBLIP \cite{instructblip}, and we instruction-tune the model, especially the Q-Former, with the language modeling loss to generate the response.

\subsection{Degradation-aware Subject and Manipulation Representation Learning}\label{sec:3.2}
With the help of the aforementioned instruction-tuned VQA module, UniProcessor can automatically generate the degradation-aware subject prompt for an input image.
However, it should be noted that UniProcessor also supports user-supplied subject prompts for feasible control.
With the development of large-scale vision-language models, including text-image feature alignment such as CLIP \cite{radford2021learning} and FLIP \cite{li2023scaling}, and text-aligned visual representation extraction such as BLIP \cite{li2022blip} and BLIP2 \cite{li2023blip2}, it is achievable to obtain text-image alignment features.
However, these features are not specifically tailored to serve as the guidance or control information.
To achieve text-guided unified image processing, we further devise a context control module to obtain the context control embedding for guidance, which is described in detail as follows.

\paragraph{Multimodal degradation-aware subject representation extraction.}
Only injecting text information to the image generation backbone can not well control a generation process \cite{rombach2022high,ruiz2023dreambooth,zhang2023adding}, thus many recent works have explored multimodal control methods \cite{zhang2023adding,gal2023designing,jia2023taming}.
As shown in Fig. \ref{fig:2_overview} (a)-(ii), we adopt the BLIP2 \cite{li2023blip2} to acquire multimodal degradation-aware subject representation.
Specifically, the input image is first encoded with a frozen pre-trained CLIP image encoder \cite{radford2021learning}, and then passed through a multimodal Q-Former to interact with the learnable queries and subject prompt.
The Q-Former produces a degradation-aware subject visual representation aligned to the subject text prompt, which is used to generate context control embedding in the next step.

\paragraph{Context-aware manipulation representation extraction.}
To achieve context-aware manipulation, the output of the above multimodal encoder is transformed using a feed-forward network containing two fully connected layers with a GELU activation in-between to obtain the subject prompt embedding, which is conformed to the input format of the text encoder.
The subject prompt embedding is then appended to the manipulation prompt with the template ``[manipulation prmpt], the [subject prompt] is [subject prompt embedding]'' to obtain a soft visual subject manipulation prompt.
Finally, the combined manipulation and subject embeddings are fed into a CLIP text encoder \cite{radford2021learning} to produce the context control embedding, which serves as the guidance information for the processor backbone to achieve controllable generation.

\begin{figure*}[t]\centering
\vspace{-0.8em}
\includegraphics[width=0.95\linewidth]{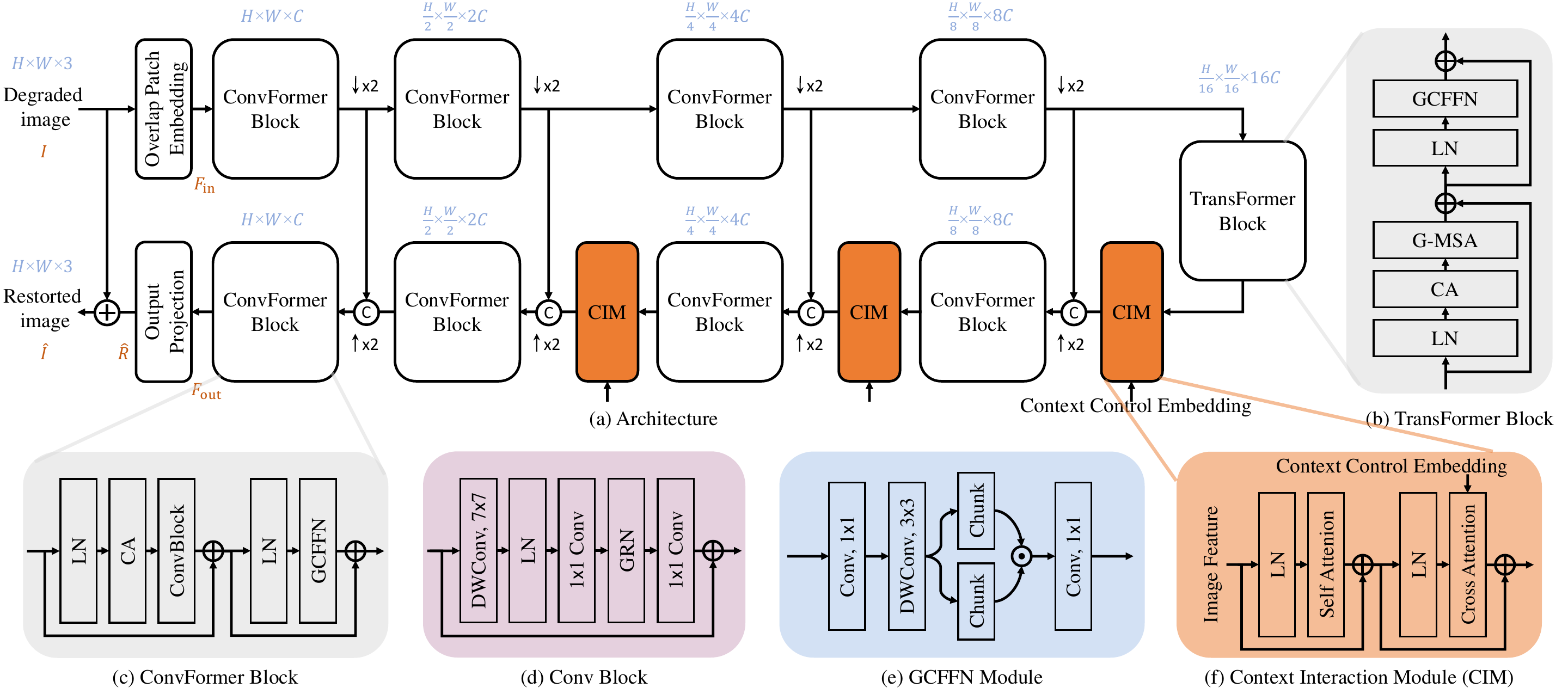}
\vspace{-0.8em}
\caption{An overview of the Processor backbone. (a) The architecture of the Processor backbone. (b) The illustration of a Transformer block. (c) The illustration of a ConvFormer block. (d) The illustration of the ConvBlock. (e) The illustration of the Gated Conv Feed-Forward Network (GCFFN). (f) The demonstration of the Context Interaction Module (CIM). LN indicates a LayerNorm layer. CA is a channel-attention layer. G-MSA represents the global multi-head self-attention. GRN means the global response normalization.
}
\label{fig:3_uniprocessor_backbone}
\vspace{-1.5em}
\end{figure*}

\vspace{-1em}
\subsection{UniProcessor with Context Control}
\vspace{-1em}

\paragraph{Processor pipeline.}
Fig. \ref{fig:3_uniprocessor_backbone}-(a) demonstrates the architecture of the processor backbone of UniProcessor. Our UniProcessor follows the design principles of encoder-decoder with skip connections, similar to UNet \cite{ronneberger2015unet}. 
For an input degraded image $\textbf{I}$, UniProcessor first uses a $3 \times 3$ convolutional layer to extract low-level feature embeddings $\textbf{F}_\text{in} \in \mathbb{R}^{H \times W \times C}$.
Next, these shallow feature maps $\textbf{F}_\text{in}$ are passed through a 5-level encoder-decoder network, then output feature maps $\textbf{F}_\text{out} \in \mathbb{R}^{H \times W \times C}$. 
Each stage of the encoder-decoder contains multiple ConvFormer or TransFormer blocks, with the number of blocks gradually increasing from the shallow level to the deep level to maintain computational efficiency.
The context interaction module (CIM) is injected into three decoding layers to guide and control the image processing procedure.
Fig. \ref{fig:3_uniprocessor_backbone}-(a) shows the feature dimensions of each level.
The pixel-unshuffle and pixel-shuffle \cite{shi2016real} methods are adopted for the down-sampling process and the up-sampling process, respectively.
We finally refine the output feature maps $\textbf{F}_\text{out}$ from the encoder-decoder network with a $3 \times 3$ convolutional layer to get the estimated residual map $\hat{\textbf{R}} \in \mathbb{R}^{H \times W \times 3}$, and obtain the restored image by $\hat{\textbf{I}} = \textbf{I} + \hat{\textbf{R}}$.
The UniProcessor is optimized using the $L_1$ loss: {\small $\mathcal{L}= || \hat{\textbf{I}}-\textbf{I}' ||$, where $\textbf{I}'$} is the ground-truth image.

\paragraph{ConvFormer and TransFormer blocks.}
Transformer \cite{vaswani2017attention,dosovitskiy2021vit} has been successfully applied to the image processing tasks \cite{chen2021IPT}.
Due to the huge computational costs of transformer architecture when applied among global image pixels, many image processing methods have been proposed to use transformer architecture in the local context or channel dimension \cite{liang2021swinir,wang2021uformer,zamir2022restormer,duan2023masked}.
As shown in Fig. \ref{fig:3_uniprocessor_backbone} (a), our UniProcessor is a hybrid architecture, which applies ConvFormer blocks (Fig. \ref{fig:3_uniprocessor_backbone} (c)) to perform local context processing and adopts TransFormer blocks (Fig. \ref{fig:3_uniprocessor_backbone} (b)) to execute global context learning.
Both the ConvFormer block and the TransFormer block contain two parts, including an attention part and a feed-forward part.
The attention part in the TransFormer block contains a channel-attention (CA) module and a global multi-head self-attention (G-MSA) module, and the feed-forward part contains a gated convolutional feed-forward network (GCFFN).
The ConvFormer block is similar to the TransFormer block, but uses a ConvBlock to perform local context awareness rather than the global context perception.
We adopt the ConvNext v2 block as the ConvBlock as shown in Fig. \ref{fig:3_uniprocessor_backbone} (d).

\begin{table*}[t]
\begin{center}
\caption{Comparison results for \textbf{30 degrations} with heavy level on the CBSD68 dataset \cite{martin2001database_bsd}.
Our model outperforms other state-of-the-art models for almost all degradation types in terms of the three most commonly used evaluation metrics, \textit{i.e.}, PSNR $\uparrow$, SSIM $\uparrow$ \cite{wang2004image}, and LPIPS $\downarrow$ \cite{zhang2018unreasonable}. The best results are colored in \textcolor{red}{red} and the second-best results are colored in \textcolor{blue}{blue}. The distortion levels in this table and more results for other distortion levels can be found in \textbf{supplemental files}.
}
\label{tab:benckmark}
\vspace{-1.5em}
\setlength{\tabcolsep}{0.45em}
\renewcommand{\arraystretch}{1.1}
\scalebox{0.38}{
\begin{tabular}{l| c | c | c | c | c | c | c | c | c }
\toprule[0.15em]
 & DRUNet \cite{zhang2021DPIR} & MPRNet \cite{Zamir_2021_CVPR_mprnet} & AirNet \cite{li2022all} & TAPE \cite{liu2022tape} & SwinIR \cite{liang2021swinir} & Uformer \cite{wang2021uformer} & Restormer \cite{zamir2022restormer} & PromptIR \cite{potlapalli2023promptir} & UniProcessor (Ours) \\
Degradation & {\fontsize{7.5}{8}\selectfont PSNR / SSIM / LPIPS} & {\fontsize{7.5}{8}\selectfont PSNR / SSIM / LPIPS} & {\fontsize{7.5}{8}\selectfont PSNR / SSIM / LPIPS} & {\fontsize{7.5}{8}\selectfont PSNR / SSIM / LPIPS} & {\fontsize{7.5}{8}\selectfont PSNR / SSIM / LPIPS} & {\fontsize{7.5}{8}\selectfont PSNR / SSIM / LPIPS} & {\fontsize{7.5}{8}\selectfont PSNR / SSIM / LPIPS} & {\fontsize{7.5}{8}\selectfont PSNR / SSIM / LPIPS} & {\fontsize{7.5}{8}\selectfont PSNR / SSIM / LPIPS} \\
\midrule[0.15em]
JPEG comp.	&	25.58	/	0.718	/	0.395	&	25.38	/	0.718	/	0.413	&	24.90	/	0.709	/	0.407	&	25.21	/	0.714	/	0.400	& 24.88	/	0.709	/	0.438	&	25.19	/	0.722	/	0.378	&	25.74	/	0.729	/	0.374	&	 \textcolor{blue}{25.85}	/	 \textcolor{blue}{0.731}	/	\textcolor{blue}{0.367}	&	\textcolor{red}{26.03}	/	\textcolor{red}{0.737}	/	 \textcolor{red}{0.367}	\\
Gauss. blur	&	23.39	/	0.578	/	0.580	&	23.29	/	0.573	/	0.597	&	22.46	/	0.528	/	0.668	&	22.56	/	0.532	/	0.624	& 22.70	/	0.539	/	0.648	&	22.99	/	0.556	/	0.617	&	24.17	/	0.621	/	0.528	&	 \textcolor{blue}{24.37}	/	 \textcolor{blue}{0.635}	/	 \textcolor{blue}{0.517}	&	\textcolor{red}{24.64}	/	\textcolor{red}{0.647}	/	\textcolor{red}{0.493}	\\
Lens blur	&	24.08	/	0.633	/	0.447	&	24.22	/	0.646	/	0.387	&	22.25	/	0.500	/	0.616	&	22.15	/	0.491	/	0.561	& 22.19	/	0.493	/	0.595	&	23.24	/	0.575	/	0.545	&	 \textcolor{blue}{26.39}	/	 \textcolor{blue}{0.759}	/	 \textcolor{blue}{0.278}	&	26.16	/	0.757	/	0.286	&	\textcolor{red}{27.35}	/	\textcolor{red}{0.798}	/	\textcolor{red}{0.212}	\\
Motion blur	&	22.09	/	0.548	/	0.537	&	21.78	/	0.532	/	0.555	&	20.96	/	0.485	/	0.587	&	21.12	/	0.499	/	0.592	& 21.09	/	0.498	/	0.602	&	21.75	/	0.528	/	0.562	&	 \textcolor{blue}{24.81}	/	 \textcolor{blue}{0.720}	/	 \textcolor{blue}{0.306}	&	24.61	/	0.700	/	0.378	&	\textcolor{red}{25.94}	/	\textcolor{red}{0.761}	/	\textcolor{red}{0.270}	\\
																																											
Color diffuse	&	21.05	/	0.862	/	0.223	&	23.15	/	0.900	/	0.174	&	20.45	/	0.870	/	0.215	&	21.95	/	0.876	/	0.197	& 21.64	/	0.887	/	0.184	&	21.92	/	0.890	/	0.187	&	23.78	/	0.910	/	0.158	&	 \textcolor{blue}{24.42}	/	 \textcolor{blue}{0.909}	/	 \textcolor{blue}{0.154}	&	\textcolor{red}{26.03}	/	\textcolor{red}{0.922}	/	\textcolor{red}{0.140}	\\
Color shift	&	34.59	/	0.986	/	0.055	&	36.71	/	0.993	/	0.034	&	36.00	/	0.991	/	0.039	&	35.77	/	0.991	/	0.038	& 34.84	/	0.991	/	0.040	&	36.16	/	0.991	/	0.035	&	 \textcolor{blue}{39.29}	/	 \textcolor{blue}{0.995}	/	 \textcolor{blue}{0.020}	&	38.68	/	0.995	/	0.024	&	\textcolor{red}{41.29}	/	\textcolor{red}{0.996}	/	\textcolor{red}{0.014}	\\
Color saturate	&	17.41	/	0.881	/	0.288	&	17.94	/	0.890	/	0.275	&	17.12	/	0.880	/	0.291	&	20.66	/	0.881	/	0.264	& 19.59	/	0.897	/	0.243	&	19.37	/	0.903	/	0.239	&	26.04	/	0.944	/	0.101	&	 \textcolor{blue}{27.21}	/	 \textcolor{blue}{0.951}	/	 \textcolor{blue}{0.084}	&	\textcolor{red}{33.15}	/	\textcolor{red}{0.978}	/	\textcolor{red}{0.024}	\\
Color saturate2	&	23.44	/	0.881	/	0.152	&	25.27	/	0.914	/	0.113	&	22.24	/	0.883	/	0.167	&	24.19	/	0.893	/	0.142	& 23.97	/	0.905	/	0.126	&	23.18	/	0.901	/	0.126	&	25.21	/	0.915	/	0.107	&	 \textcolor{blue}{25.88}	/	 \textcolor{blue}{0.919}	/	 \textcolor{blue}{0.099}	&	\textcolor{red}{27.39}	/	\textcolor{red}{0.929}	/	\textcolor{red}{0.084}	\\
Gauss. noise	&	26.40	/	0.723	/	0.297	&	 \textcolor{blue}{26.48}	/	0.723	/	0.301	&	26.34	/	0.721	/	0.269	&	25.87	/	0.677	/	0.303	& 26.42	/	0.723	/	0.285	&	26.32	/	0.732	/	\textcolor{red}{0.228}	&	26.45	/	0.739	/	0.258	&	26.34	/	 \textcolor{blue}{0.747}	/	0.238	&	\textcolor{red}{26.51}	/	\textcolor{red}{0.766}	/	 \textcolor{blue}{0.230}	\\
GN (ycbcr)	&	29.52	/	0.839	/	0.164	&	29.68	/	0.841	/	0.166	&	29.52	/	0.841	/	0.144	&	29.15	/	0.818	/	0.164	& 29.49	/	0.834	/	0.166	&	29.44	/	0.845	/	 \textcolor{blue}{0.128}	&	29.93	/	0.853	/	0.136	&	 \textcolor{blue}{29.93}	/	 \textcolor{blue}{0.855}	/	0.134	&	\textcolor{red}{30.40}	/	\textcolor{red}{0.868}	/	\textcolor{red}{0.125}	\\
Impulse noise	&	39.08	/	0.985	/	0.013	&	40.38	/	0.988	/	0.009	&	36.49	/	0.971	/	0.030	&	38.76	/	0.982	/	0.018	& 40.44	/	0.986	/	0.012	&	37.71	/	0.982	/	0.015	&	 \textcolor{blue}{42.28}	/	0.992	/	 \textcolor{blue}{0.006}	&	42.06	/	 \textcolor{blue}{0.992}	/	0.006	&	\textcolor{red}{42.74}	/	\textcolor{red}{0.993}	/	\textcolor{red}{0.003}	\\
Multipli. noise	&	29.66	/	0.867	/	0.146	&	30.38	/	0.877	/	0.137	&	30.02	/	0.873	/	0.127	&	29.74	/	0.856	/	0.135	& 30.23	/	0.872	/	0.136	&	30.28	/	0.883	/	\textcolor{red}{0.099}	&	 \textcolor{blue}{30.62}	/	0.885	/	0.114	&	30.55	/	 \textcolor{blue}{0.887}	/	0.113	&	\textcolor{red}{31.25}	/	\textcolor{red}{0.901}	/	 \textcolor{blue}{0.102}	\\
Denoise	&	 \textcolor{blue}{24.78}	/	0.657	/	0.553	&	24.73	/	0.648	/	0.568	&	24.18	/	0.616	/	0.626	&	24.14	/	0.623	/	0.593	& 24.05	/	0.613	/	0.636	&	24.72	/	0.657	/	0.531	&	24.76	/	0.646	/	0.586	&	24.75	/	 \textcolor{blue}{0.658}	/	 \textcolor{blue}{0.519}	&	\textcolor{red}{25.15}	/	\textcolor{red}{0.673}	/	\textcolor{red}{0.475}	\\
Over bright	&	15.01	/	0.742	/	0.269	&	18.01	/	0.800	/	0.228	&	14.06	/	0.683	/	0.313	&	16.90	/	0.757	/	0.265	& 19.08	/	0.842	/	0.178	&	13.94	/	0.770	/	0.226	&	20.63	/	0.877	/	0.149	&	 \textcolor{blue}{21.64}	/	 \textcolor{blue}{0.882}	/	 \textcolor{blue}{0.141}	&	\textcolor{red}{23.76}	/	\textcolor{red}{0.905}	/	\textcolor{red}{0.112}	\\
Low-light	&	14.67	/	0.613	/	0.286	&	19.60	/	0.751	/	0.232	&	12.59	/	0.466	/	0.363	&	17.86	/	0.706	/	0.260	& 18.90	/	0.744	/	0.240	&	21.27	/	0.801	/	0.175	&	20.37	/	0.787	/	0.179	&	 \textcolor{blue}{24.07}	/	 \textcolor{blue}{0.837}	/	 \textcolor{blue}{0.154}	&	\textcolor{red}{24.23}	/	\textcolor{red}{0.846}	/	\textcolor{red}{0.138}	\\
Mean shift	&	19.13	/	0.866	/	0.070	&	22.74	/	0.913	/	0.052	&	16.91	/	0.777	/	0.072	&	18.51	/	0.859	/	0.089	& 19.28	/	0.862	/	0.093	&	23.26	/	0.926	/	0.041	&	23.26	/	0.915	/	0.035	&	 \textcolor{blue}{24.41}	/	 \textcolor{blue}{0.927}	/	 \textcolor{blue}{0.031}	&	\textcolor{red}{27.99}	/	\textcolor{red}{0.947}	/	\textcolor{red}{0.023}	\\
Bicubic resize/SR	&	20.98	/	0.462	/	0.734	&	20.84	/	0.455	/	0.740	&	20.73	/	0.450	/	0.766	&	20.72	/	0.449	/	0.748	& 20.74	/	0.449	/	0.775	&	20.94	/	0.461	/	0.730	&	21.12	/	0.468	/	 \textcolor{blue}{0.709}	&	 \textcolor{blue}{21.17}	/	 \textcolor{blue}{0.469}	/	0.723	&	\textcolor{red}{21.25}	/	\textcolor{red}{0.474}	/	\textcolor{red}{0.697}	\\
Bilinear resize/SR	&	20.83	/	0.457	/	0.737	&	20.68	/	0.451	/	0.736	&	20.25	/	0.437	/	0.791	&	20.33	/	0.439	/	0.757	& 20.47	/	0.442	/	0.767	&	20.75	/	0.457	/	0.726	&	21.05	/	0.466	/	 \textcolor{blue}{0.706}	&	 \textcolor{blue}{21.10}	/	 \textcolor{blue}{0.468}	/	0.725	&	\textcolor{red}{21.21}	/	\textcolor{red}{0.472}	/	\textcolor{red}{0.701}	\\
Nearest resize/SR	&	22.08	/	0.574	/	0.481	&	21.95	/	0.572	/	0.508	&	21.73	/	0.569	/	0.475	&	21.81	/	0.563	/	0.534	& 21.80	/	0.563	/	0.526	&	22.06	/	0.573	/	0.493	&	22.07	/	 \textcolor{blue}{0.579}	/	0.483	&	 \textcolor{blue}{22.16}	/	0.579	/	 \textcolor{blue}{0.478}	&	\textcolor{red}{22.32}	/	\textcolor{red}{0.585}	/	\textcolor{red}{0.452}	\\
Lanczos resize/SR	&	21.01	/	0.461	/	0.749	&	20.91	/	0.456	/	0.748	&	20.85	/	0.453	/	0.763	&	20.86	/	0.452	/	0.745	& 20.87	/	0.452	/	0.774	&	20.99	/	0.461	/	0.742	&	21.15	/	0.468	/	 \textcolor{blue}{0.714}	&	 \textcolor{blue}{21.17}	/	 \textcolor{blue}{0.468}	/	0.728	&	\textcolor{red}{21.27}	/	\textcolor{red}{0.473}	/	\textcolor{red}{0.702}	\\
Sharpening	&	25.24	/	0.887	/	0.123	&	25.03	/	0.884	/	0.123	&	25.30	/	0.874	/	0.144	&	25.10	/	0.866	/	0.146	& 25.39	/	0.896	/	0.106	&	25.11	/	0.881	/	0.128	&	25.69	/	0.896	/	0.118	&	 \textcolor{blue}{26.65}	/	 \textcolor{blue}{0.917}	/	 \textcolor{blue}{0.088}	&	\textcolor{red}{27.60}	/	\textcolor{red}{0.930}	/	\textcolor{red}{0.072}	\\
Contrast imbal.	&	21.97	/	0.889	/	0.124	&	22.01	/	0.887	/	0.122	&	21.92	/	0.888	/	0.122	&	26.52	/	0.936	/	0.088	& 23.66	/	0.872	/	0.160	&	22.67	/	0.902	/	0.117	&	30.00	/	0.976	/	0.043	&	 \textcolor{blue}{33.13}	/	 \textcolor{blue}{0.982}	/	 \textcolor{blue}{0.023}	&	\textcolor{red}{40.43}	/	\textcolor{red}{0.994}	/	\textcolor{red}{0.004}	\\
Color block	&	30.83	/	0.958	/	0.072	&	30.55	/	0.959	/	0.074	&	24.80	/	0.936	/	0.139	&	28.20	/	0.951	/	0.095	& 30.74	/	0.959	/	0.065	&	32.08	/	0.965	/	 \textcolor{blue}{0.057}	&	32.01	/	0.964	/	0.061	&	 \textcolor{blue}{32.30}	/	 \textcolor{blue}{0.966}	/	0.058	&	\textcolor{red}{33.09}	/	\textcolor{red}{0.969}	/	\textcolor{red}{0.051}	\\
Pixelate	&	24.22	/	0.697	/	0.331	&	24.00	/	0.696	/	0.334	&	23.99	/	0.695	/	0.348	&	23.85	/	0.689	/	0.383	& 23.91	/	0.692	/	0.372	&	23.98	/	0.698	/	0.339	&	24.14	/	0.702	/	0.332	&	 \textcolor{blue}{24.22}	/	 \textcolor{blue}{0.703}	/	 \textcolor{blue}{0.327}	&	\textcolor{red}{24.41}	/	\textcolor{red}{0.710}	/	\textcolor{red}{0.306}	\\
Discontinuous	&	26.76	/	0.897	/	0.074	&	26.37	/	0.898	/	 \textcolor{blue}{0.061}	&	25.16	/	0.886	/	0.076	&	25.03	/	0.884	/	0.088	& 25.14	/	0.886	/	0.080	&	25.00	/	0.887	/	0.077	&	 \textcolor{blue}{29.04}	/	 \textcolor{blue}{0.920}	/	0.068	&	27.33	/	0.907	/	\textcolor{red}{0.057}	&	\textcolor{red}{29.65}	/	\textcolor{red}{0.929}	/	\textcolor{red}{0.062}	\\
Jitter	&	23.85	/	0.634	/	\textcolor{red}{0.387}	&	23.91	/	0.647	/	0.404	&	23.55	/	0.625	/	0.373	&	23.27	/	0.618	/	0.418	& 23.59	/	0.629	/	0.425	&	23.84	/	0.638	/	 \textcolor{blue}{0.390}	&	24.07	/	0.652	/	0.405	&	 \textcolor{blue}{24.13}	/	 \textcolor{blue}{0.652}	/	0.412	&	\textcolor{red}{24.29}	/	\textcolor{red}{0.661}	/	0.426	\\
Mosaic	&	36.46	/	0.984	/	0.017	&	35.66	/	0.972	/	0.017	&	34.70	/	0.978	/	0.025	&	36.45	/	0.977	/	0.025	& 35.07	/	0.976	/	0.021	&	11.45	/	0.423	/	0.428	&	38.18	/	0.983	/	 \textcolor{blue}{0.016}	&	 \textcolor{blue}{38.21}	/	 \textcolor{blue}{0.984}	/	0.018	&	\textcolor{red}{40.62}	/	\textcolor{red}{0.990}	/	\textcolor{red}{0.009}	\\
Irregular mask	&	27.43	/	0.885	/	0.161	&	28.29	/	0.889	/	0.155	&	26.74	/	0.868	/	0.177	&	26.84	/	0.875	/	0.167	& 27.17	/	0.882	/	0.154	&	27.74	/	0.894	/	 \textcolor{blue}{0.143}	&	28.57	/	0.891	/	0.153	&	 \textcolor{blue}{29.38}	/	 \textcolor{blue}{0.894}	/	0.150	&	\textcolor{red}{29.97}	/	\textcolor{red}{0.900}	/	\textcolor{red}{0.138}	\\
Block mask	&	27.72	/	0.916	/	0.129	&	29.10	/	0.920	/	0.123	&	27.27	/	0.906	/	0.134	&	26.35	/	0.905	/	0.140	& 27.05	/	0.913	/	0.124	&	29.22	/	0.923	/	 \textcolor{blue}{0.115}	&	28.69	/	0.920	/	0.122	&	 \textcolor{blue}{30.67}	/	 \textcolor{blue}{0.923}	/	0.119	&	\textcolor{red}{32.07}	/	\textcolor{red}{0.926}	/	\textcolor{red}{0.114}	\\
Rain streak	&	24.34	/	0.769	/	0.203	&	25.26	/	0.808	/	0.173	&	17.17	/	0.668	/	0.300	&	20.32	/	0.756	/	0.216	& 23.38	/	0.804	/	0.170	&	16.39	/	0.702	/	0.254	&	26.89	/	0.847	/	0.123	&	 \textcolor{blue}{27.80}	/	 \textcolor{blue}{0.867}	/	 \textcolor{blue}{0.099}	&	\textcolor{red}{28.67}	/	\textcolor{red}{0.890}	/	\textcolor{red}{0.082}	\\
Snow streak	&	22.89	/	0.644	/	0.382	&	24.14	/	0.716	/	0.318	&	24.64	/	0.745	/	0.284	&	25.19	/	0.781	/	0.233	& 25.64	/	0.806	/	0.215	&	 \textcolor{blue}{26.94}	/	 \textcolor{blue}{0.830}	/	 \textcolor{blue}{0.170}	&	21.47	/	0.620	/	0.398	&	24.67	/	0.723	/	0.294	&	\textcolor{red}{30.20}	/	\textcolor{red}{0.885}	/	\textcolor{red}{0.116}	\\
																																											
Clean image	&	41.23	/	0.990	/	0.009	&	53.36	/	0.999	/	0.001	&	47.76	/	0.996	/	0.003	&	41.98	/	0.994	/	0.011	& 39.31	/	0.995	/	0.006	&	45.20	/	0.996	/	0.003	&	49.17	/	0.997	/	0.002	&	\textcolor{blue}{62.24}	/	 \textcolor{blue}{0.998}	/	 \textcolor{blue}{0.001}	&	\textcolor{red}{80.16}	/	\textcolor{red}{1.000}	/	\textcolor{red}{0.000}	\\

\midrule[0.15em]
                                           
Average	&	25.24	/	0.765	/	0.287	&	26.31	/	0.779	/	0.277	&	24.47	/	0.743	/	0.308	&	25.23	/	0.759	/	0.295	& 25.40	/	0.769	/	0.293	&	24.85	/	0.761	/	0.283	&	27.41	/	0.801	/	0.243	&	\textcolor{blue}{28.35}	/	\textcolor{blue}{0.809}	/	\textcolor{blue}{0.236}	&	\textcolor{red}{\textbf{30.35}}	/	\textcolor{red}{\textbf{0.827}}	/	\textcolor{red}{\textbf{0.211}}	\\

\bottomrule[0.15em]
\end{tabular}}
\end{center}
\vspace{-2.5em}
\end{table*}

\paragraph{Channel-attention module and gated convolutional feed-forward module.}
UniProcessor adopts a channel-attention module \cite{chen2022nafnet} to perform channel-wise feature refinement.
Given an input tensor $\textbf{X}$, the output of the CA layer can be formulated as: $\text{CA}(\textbf{X})=\textbf{X}*\text{MLP}(\text{Avg}(\textbf{X}))$, where Avg is an average pooling layer, MLP is a multilayer perceptron, $*$ indicates a channel-wise product operation.
As shown in Fig. \ref{fig:3_uniprocessor_backbone} (e), UniProcessor adopts a gated convolutional feed-forward network (GCFFN) \cite{dauphin2017language,shazeer2020glu,zamir2022restormer, duan2023masked} as the feed-forward network.
Give an input tensor $\textbf{X}$, the GCFFN process can be formulated as: $\textbf{X}_1 = W_d^1(W_p^1(\textbf{X}))$, $\textbf{X}_2 = W_d^2(W_p^2(\textbf{X}))$, $\hat{\textbf{X}} = W_p^3(\phi(\textbf{X}_1)\odot \textbf{X}_2)$, where $W_p$ represents a $1 \times 1$ point-wise convolution, $W_d$ indicates a $3 \times 3$ depth-wise convolution, $\phi$ is the GELU operation, $\odot$ denotes element-wise multiplication.

\paragraph{Context Interaction Module.}
The highlight of our UniProcessor is to achieve text-guided image processing, which is mainly accomplished by the context interaction module (CIM).
In Section \ref{sec:3.2}, we have obtained the degradation-aware context control embedding, which is represented as $\textbf{E}$ here.
The primary goal of the context interaction module is to enable interaction between the input feature $\textbf{F}$ and the context control embedding $\textbf{E}$.
As shown in Fig. \ref{fig:3_uniprocessor_backbone} (f), the CIM contains a self-attention block and a cross-attention block with LayerNorm (LN) layers before. The overall process of the CIM is:
\begin{equation}
\vspace{-0.3em}
    \textbf{F}' = \text{Self-Attn}(\text{LN}(\textbf{F})) + \textbf{F},
\end{equation}
\vspace{-1.4em}
\begin{equation}
    \textbf{F}'' = \text{Cross-Attn}(\text{LN}(\textbf{F}'), \textbf{E}) + \textbf{F}',
\end{equation}
where LN, Self-Attn, Cross-Attn represent layernorm, self-attention, and cross-attention, respectively, and $\textbf{F}'$ is the middle feature, $\textbf{F}''$ is the output integrated feature of the CIM module.
CIM is a plug-in module, which controls the processing procedure by interacting image features and contextual control embeddings through cross-attention layers.

\begin{figure*}[!t]
\vspace{-0.5em}
\begin{center}
\setlength{\tabcolsep}{0.15em}
\scalebox{0.65}{
\begin{tabular}[b]{c c c c c c c c}
\adjincludegraphics[trim={ 0 0 0 {.20\height} },clip,width=.185\textwidth,valign=t]{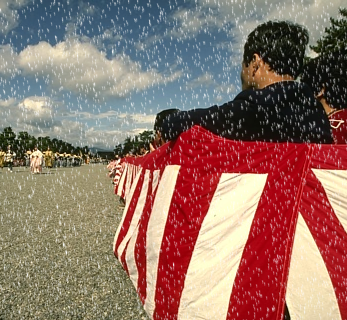} &  
\adjincludegraphics[trim={ 0 0 0 {.20\height} },clip,width=.185\textwidth,valign=t]{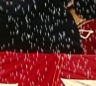} &  
\adjincludegraphics[trim={ 0 0 0 {.20\height} },clip,width=.185\textwidth,valign=t]{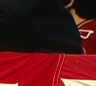} &  
\adjincludegraphics[trim={ 0 0 0 {.20\height} },clip,width=.185\textwidth,valign=t]{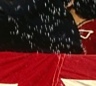} &   
\adjincludegraphics[trim={ 0 0 0 {.20\height} },clip,width=.185\textwidth,valign=t]{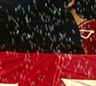} &   
\adjincludegraphics[trim={ 0 0 0 {.20\height} },clip,width=.185\textwidth,valign=t]{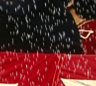} &  
\adjincludegraphics[trim={ 0 0 0 {.20\height} },clip,width=.185\textwidth,valign=t]{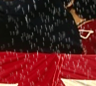} &  
\adjincludegraphics[trim={ 0 0 0 {.20\height} },clip,width=.185\textwidth,valign=t]{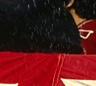} 
\\
\small~Snowy~image & \small~Snowy~patch & \small~Reference & \small~MPRNet~\cite{Zamir_2021_CVPR_mprnet} & \small~SwinIR~\cite{liang2021swinir} & \small~Restormer~\cite{zamir2022restormer}  & \small~PromptIR~\cite{potlapalli2023promptir} & \small~\textbf{UniProcessor}
\\

\adjincludegraphics[trim={ 0 0 0 {.20\height} },clip,width=.185\textwidth,valign=t]{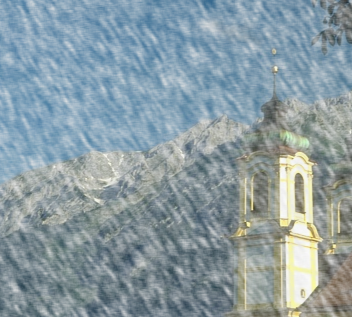} &  
\adjincludegraphics[trim={ 0 0 0 {.20\height} },clip,width=.185\textwidth,valign=t]{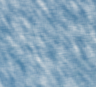} &  
\adjincludegraphics[trim={ 0 0 0 {.20\height} },clip,width=.185\textwidth,valign=t]{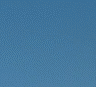} &  
\adjincludegraphics[trim={ 0 0 0 {.20\height} },clip,width=.185\textwidth,valign=t]{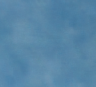} &   
\adjincludegraphics[trim={ 0 0 0 {.20\height} },clip,width=.185\textwidth,valign=t]{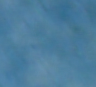} &   
\adjincludegraphics[trim={ 0 0 0 {.20\height} },clip,width=.185\textwidth,valign=t]{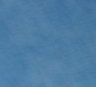} &  
\adjincludegraphics[trim={ 0 0 0 {.20\height} },clip,width=.185\textwidth,valign=t]{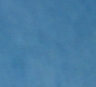} &  
\adjincludegraphics[trim={ 0 0 0 {.20\height} },clip,width=.185\textwidth,valign=t]{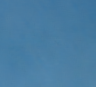} 
\\
\small~Rainy~image & \small~Rainy~patch & \small~Reference & \small~MPRNet~\cite{Zamir_2021_CVPR_mprnet} & \small~SwinIR~\cite{liang2021swinir} & \small~Restormer~\cite{zamir2022restormer}  & \small~PromptIR~\cite{potlapalli2023promptir} & \small~\textbf{UniProcessor}
\\

\adjincludegraphics[trim={ 0 0 0 {.20\height} },clip,width=.185\textwidth,valign=t]{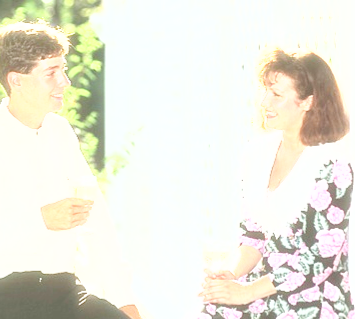} &  
\adjincludegraphics[trim={ 0 0 0 {.20\height} },clip,width=.185\textwidth,valign=t]{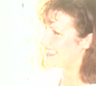} &  
\adjincludegraphics[trim={ 0 0 0 {.20\height} },clip,width=.185\textwidth,valign=t]{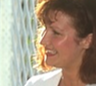} &  
\adjincludegraphics[trim={ 0 0 0 {.20\height} },clip,width=.185\textwidth,valign=t]{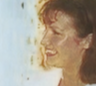} &   
\adjincludegraphics[trim={ 0 0 0 {.20\height} },clip,width=.185\textwidth,valign=t]{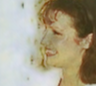} &   
\adjincludegraphics[trim={ 0 0 0 {.20\height} },clip,width=.185\textwidth,valign=t]{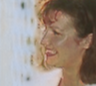} &  
\adjincludegraphics[trim={ 0 0 0 {.20\height} },clip,width=.185\textwidth,valign=t]{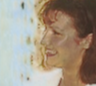} &  
\adjincludegraphics[trim={ 0 0 0 {.20\height} },clip,width=.185\textwidth,valign=t]{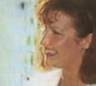} 
\\
\small~Over~bright & \small~Over~bright & \small~Reference & \small~MPRNet~\cite{Zamir_2021_CVPR_mprnet} & \small~SwinIR~\cite{liang2021swinir} & \small~Restormer~\cite{zamir2022restormer}  & \small~PromptIR~\cite{potlapalli2023promptir} & \small~\textbf{UniProcessor}
\\

\adjincludegraphics[trim={ 0 0 0 {.20\height} },clip,width=.185\textwidth,valign=t]{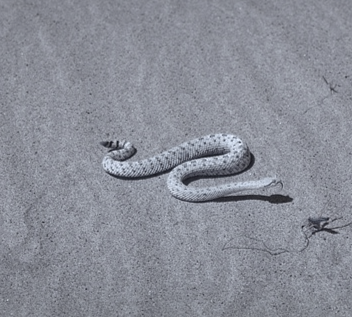} &  
\adjincludegraphics[trim={ 0 0 0 {.20\height} },clip,width=.185\textwidth,valign=t]{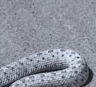} &  
\adjincludegraphics[trim={ 0 0 0 {.20\height} },clip,width=.185\textwidth,valign=t]{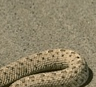} &  
\adjincludegraphics[trim={ 0 0 0 {.20\height} },clip,width=.185\textwidth,valign=t]{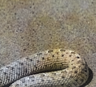} &   
\adjincludegraphics[trim={ 0 0 0 {.20\height} },clip,width=.185\textwidth,valign=t]{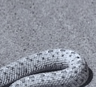} &   
\adjincludegraphics[trim={ 0 0 0 {.20\height} },clip,width=.185\textwidth,valign=t]{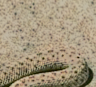} &  
\adjincludegraphics[trim={ 0 0 0 {.20\height} },clip,width=.185\textwidth,valign=t]{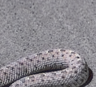} &  
\adjincludegraphics[trim={ 0 0 0 {.20\height} },clip,width=.185\textwidth,valign=t]{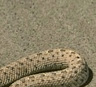} 
\\
\small~Color~Saturate & \small~Color~Saturate & \small~Reference & \small~MPRNet~\cite{Zamir_2021_CVPR_mprnet} & \small~SwinIR~\cite{liang2021swinir} & \small~Restormer~\cite{zamir2022restormer}  & \small~PromptIR~\cite{potlapalli2023promptir} & \small~\textbf{UniProcessor}
\\

\adjincludegraphics[trim={ 0 0 0 {.20\height} },clip,width=.185\textwidth,valign=t]{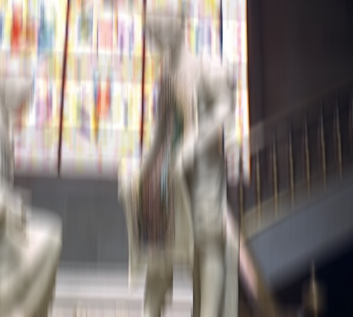} &  
\adjincludegraphics[trim={ 0 0 0 {.20\height} },clip,width=.185\textwidth,valign=t]{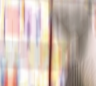} &  
\adjincludegraphics[trim={ 0 0 0 {.20\height} },clip,width=.185\textwidth,valign=t]{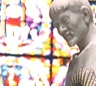} &  
\adjincludegraphics[trim={ 0 0 0 {.20\height} },clip,width=.185\textwidth,valign=t]{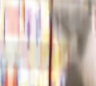} &   
\adjincludegraphics[trim={ 0 0 0 {.20\height} },clip,width=.185\textwidth,valign=t]{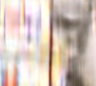} &   
\adjincludegraphics[trim={ 0 0 0 {.20\height} },clip,width=.185\textwidth,valign=t]{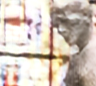} &  
\adjincludegraphics[trim={ 0 0 0 {.20\height} },clip,width=.185\textwidth,valign=t]{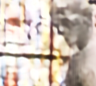} &  
\adjincludegraphics[trim={ 0 0 0 {.20\height} },clip,width=.185\textwidth,valign=t]{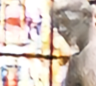} 
\\
\small~Motion-blur & \small~Motion-blur & \small~Reference & \small~MPRNet~\cite{Zamir_2021_CVPR_mprnet} & \small~SwinIR~\cite{liang2021swinir} & \small~Restormer~\cite{zamir2022restormer}  & \small~PromptIR~\cite{potlapalli2023promptir} & \small~\textbf{UniProcessor}
\\

\end{tabular}}
\end{center}
\vspace{-1.7em}
\caption{Visualization results on 5 different degradation types. UniProcessor produces more visually pleasant results.
}
\label{fig:4_visualization}
\vspace{-1.5em}
\end{figure*}

\vspace{-0.8em}
\section{Experiments}
\vspace{-0.6em}
In this section, we conduct experiments to validate the effectiveness of the proposed method in learning an all-in-one image processing model, and demonstrate the good degradation awareness and disentangling ability of our UniProcessor. Due to the page limitation, more experimental results are shown in the \textit{supplementary material}.

\vspace{-0.8em}
\subsection{Experimental Setup}
\vspace{-0.5em}

\paragraph{Implementation details.} 
To enable the degradation-aware VQA capabilities of our model, we first adopt the vision language instruction tuning strategy to tune the VQA module of UniProcessor.
Specifically, we initialize the weights of the multimodal Q-Former in the instruction-tuned module with pre-trained InstructBLIP \cite{instructblip}, which empowers the model with initial ability of complex visual scene understanding.
The frozen visual encoder is a ViT-G from EVA-CLIP \cite{fang2023eva}, and the frozen LLM used in UniProcessor is Vicuna \cite{zheng2023judging}, which is a decoder-only LLM fine-tuned from LLaMA \cite{touvron2023llama}.
We fine-tune it on the constructed low-level VQA dataset to adapt the frozen LLM to give detailed feedback for the degradations of an image.
The model is trained using the standard language modeling loss.

\noindent The image encoder in CIM is same as that in the VQA module, and the text encoder is from BLIP \cite{li2022blip}.
The architecture of our UniProcessor consists of a five-level encoder-decoder, with [4, 6, 6, 8, 8] ConvFormer or TransFormer layers from level-1 to level-4, respectively.
For CIM-plugged decoding layers, each layer contains two CIM blocks, and the total number of CIM components in UniProcessor is 6.
The UniProcessor model is trained on 224 $\times$ 224 random-cropped patches with random data augmentation methods including horizontal and vertical flips, $90^{\circ}$ rotations, \textit{etc}.
We use AdamW Optimizer to train the network for 150 epochs with a batch size of 36.
The initial learning rate is 2$e^{-4}$ and gradually reduces to 1$e^{-6}$ with cosine annealing \cite{loshchilovsgdr}.

\paragraph{Database preparation.}
We adopt a combined set including 900 images from DIV2K~\cite{agustsson2017ntire}, 2650 images from Flickr2K, 400 images from BSD500~\cite{arbelaez2010contour}, and 4744 images from WaterlooED~(WED)~\cite{ma2016waterloo}, as the training dataset, and use four datasets, including CBSD68 \cite{martin2001database_bsd}, Urban100 \cite{huang2015single_urban100}, Kodak24 \cite{kodak}, and McMaster \cite{zhang2011color_mcmaster}, as the test dataset.
We first crop the training images into $512 \times 512$ patches with a stride of $416$ to generate 71580 small patches \cite{zamir2022restormer} for training.
During the training process, we randomly crop desired patches from the $512 \times 512$ patches prepared above as training patches.
The degradations are generated on-the-fly during training process.
We first develop a distortion bank, which includes 30 common degradations with various levels  \cite{zhang2017learning,saha2023re,suvorov2022resolution} (see the \textit{supplementary material}).
During training, we randomly adjust the degradation level to cover a wide perception range for improving the generality of the proposed model.
For testing, we set three levels of degradations including heavy, middle and slight.

\vspace{-1.1em}
\subsection{Multiple Degradation All-in-one Results}
\vspace{-0.4em}
We conduct extensive experiments to evaluate the all-in-one image processing results of our proposed UniProcessor as well as six state-of-the-art image restoration methods.
These representative methods include DRUNet \cite{zhang2021DPIR}, MPRNet \cite{Zamir_2021_CVPR_mprnet}, SwinIR \cite{liang2021swinir}, Restormer \cite{zamir2022restormer}, and PromptIR \cite{potlapalli2023promptir}.
These competing methods are retrained in the experiments with their publicly released codes and following their original settings, under our data preparation setting.
All models are trained for 150 epochs using the same training set and degradation generation methods.

\paragraph{Quantitative comparison results.}
Table \ref{tab:benckmark} quantitatively demonstrates the performance results of our UniProcessor and six competing models for processing 30 severe degradations on the CBSD68 dataset  \cite{martin2001database_bsd}.
It can be observed that our UniProcessor achieves state-of-the-art performance and outperforms other models for almost all degradation types in terms of three commonly used evaluation metrics, \textit{i.e.}, PSNR, SSIM \cite{wang2004image}, and LPIPS \cite{zhang2018unreasonable}, which manifests the effectiveness of the proposed method.
Moreover, our method achieves consistent improvement but different amounts for various tasks, \textit{e.g.}, 0.2dB for JPEG but 5.5dB for snow removal compared to PromptIR, indicating saturate improvement for some tasks.
More quantitative results can be found in the \textit{supplementary material}.

\begin{SCfigure*}[][b]
\centering
\caption{
tSNE plots of the degradation embeddings in UniProcessor (ours) and the state-of-the-art model PromptIR \cite{potlapalli2023promptir}. Our results are better clustered, manifesting the effectiveness of text-induced prompt method for learning discriminative degradation context.
}
\label{fig:tsne}
\includegraphics[scale=0.32]{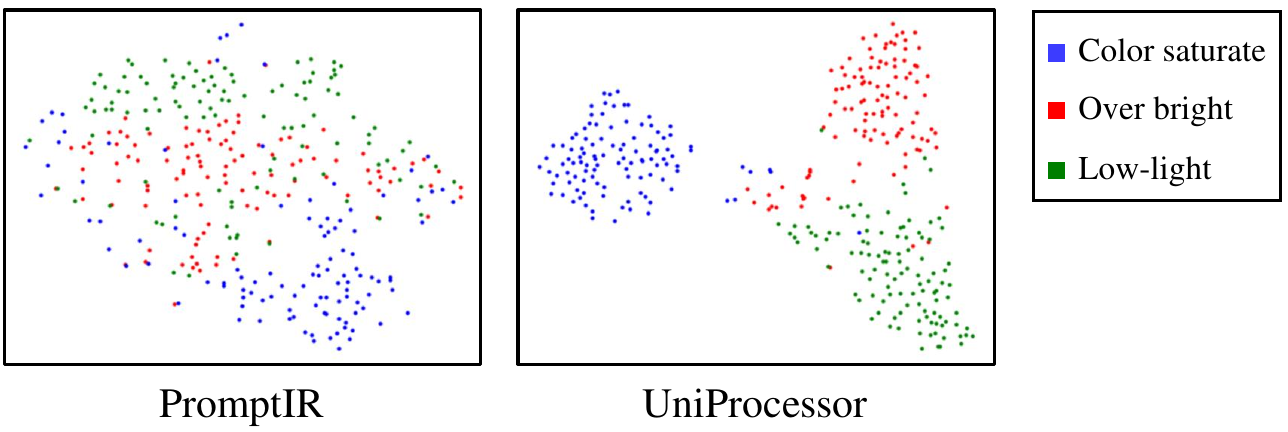}\vspace{-0.0em}
\end{SCfigure*}
\begin{figure}[t]\centering
\vspace{-0.5em}
\includegraphics[width=1\linewidth]{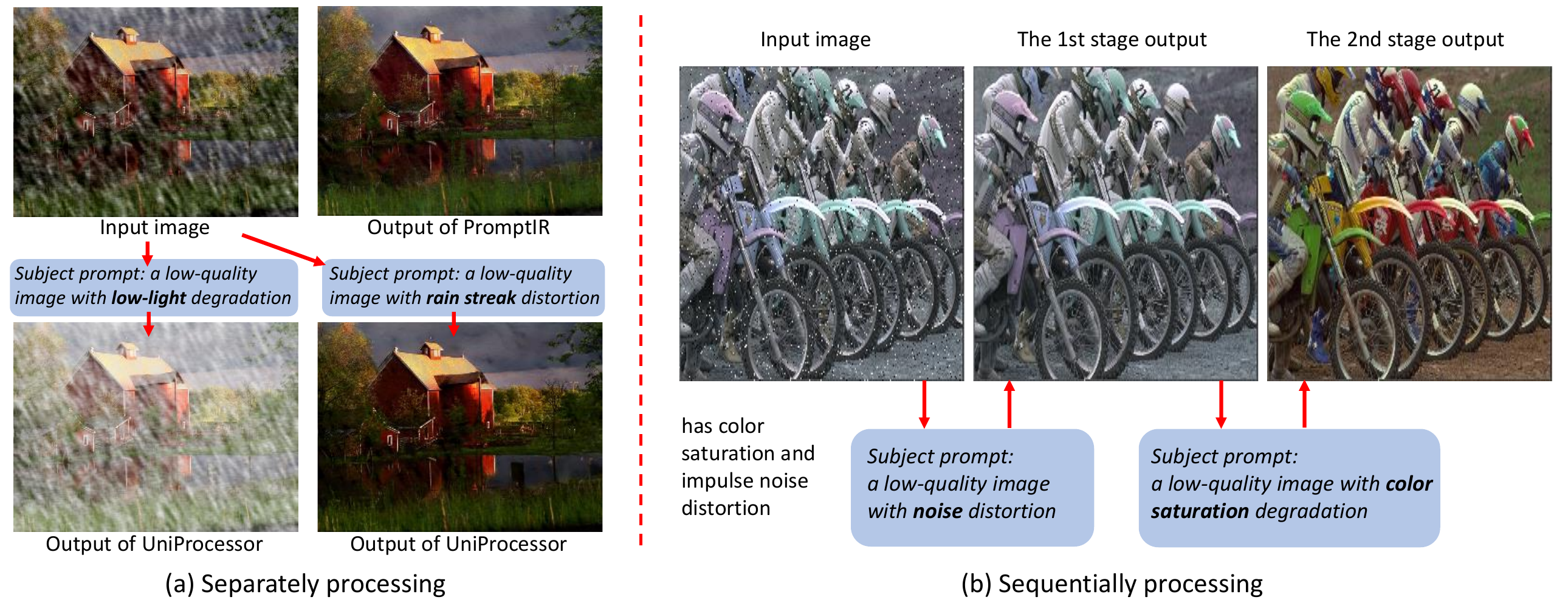}
\vspace{-1.8em}
\caption{UniProcessor can independently handle individual degradations in an image with multiple distortions through different subject prompt controls, and can gradually process the multiple distortions step by step.
}
\label{fig:5_individually_process}
\end{figure}

\paragraph{Qualitative comparison results.}
Fig. \ref{fig:4_visualization} shows the visual comparisons of the results from our UniProcessor and other state-of-the-art restoration models on 5 different degradation types.
It qualitatively demonstrates that our UniProcessor can well process these degraded inputs and restore high-quality clean images using an all-in-one model.
Moreover, compared to other competing methods, UniProcessor generates more visually-faithful results.

\paragraph{tSNE results.}
Fig. \ref{fig:tsne} shows the tSNE plots of the degradation embeddings in UniProcessor and the state-of-the-art all-in-one restoration model PromptIR \cite{potlapalli2023promptir}.
Distinct colors represent different degradation types. 
The embeddings for the three tasks are better clustered in our case, which manifests that UniProcessor can effectively learn discriminative features for recognizing the degradations.

\paragraph{Multi-degradation separately processing results.}
Due to the degradation-aware and context manipulation capabilities, our UniProcessor can well disentangle the degradations and achieve the ability to individually process a single degradation in an image with multiple distortions.
As shown in Fig. \ref{fig:5_individually_process} (a), for the input image with low-light and rain streak degradations, the PromptIR \cite{potlapalli2023promptir} model can only output one restored image, and only removes the most influential degradation, \textit{i.e.}, rain streaks.
However, for UniProcessor, with different subject prompt control, we can control the process more flexibly and generate the desired output.

\paragraph{Multi-degradation gradually processing results.}
The above experiment demonstrates that UniProcessor can well disentangle degradations and individually process them.
We further conduct an experiment to demonstrate that UniProcessor has the ability to remove multiple degradations step by step.
As shown in Fig. \ref{fig:5_individually_process} (b), for an input with color saturation and noise distortion, UniProcessor can first remove noise with the noise-related subject prompt.
The output is served as the second stage input and we process the image with the color saturation-corresponding subject prompt to obtain the final high-quality image.

\begin{table}[t]
\begin{center}
\vspace{-0.3em}
\caption{Quantitative comparison for multi-degradation separately processing and multi-degradation gradually processing. PromptIR 2-step: process an image using PromptIR twice. Degradation contains \underline{\textbf{\textit{d1}}} and \underline{\textbf{\textit{d2}}}.
The \underline{\textbf{\textit{rm d1}}}, \underline{\textbf{\textit{rm d2}}}, \underline{\textbf{\textit{rm d1+d2}}}, \underline{\textbf{\textit{rm d2+d1}}} means: remove \underline{\textbf{\textit{d1}}}, remove \underline{\textbf{\textit{d2}}}, first remove \underline{\textbf{\textit{d1}}} then \underline{\textbf{\textit{d2}}}, first \underline{\textbf{\textit{d2}}} then \underline{\textbf{\textit{d1}}}, respectively.
}
\label{tab2}
\vspace{-1.3em}
\setlength{\tabcolsep}{0.2em}
\renewcommand{\arraystretch}{0.5}
\scalebox{0.5}{
\begin{tabular}{l| c | c | c | c | c | c | c }
\toprule[0.15em]
 & {\fontsize{7.6}{8}\selectfont Restormer \cite{zamir2022restormer}} & {\fontsize{7.6}{8}\selectfont PromptIR \cite{potlapalli2023promptir}} & {\fontsize{7.6}{8}\selectfont PromptIR \cite{potlapalli2023promptir} 2-step} & {\fontsize{7.6}{8}\selectfont UniProcessor \underline{\textbf{\textit{rm d1}}}} & {\fontsize{7.6}{8}\selectfont UniProcessor \underline{\textbf{\textit{rm d2}}}} & {\fontsize{7.6}{8}\selectfont UniProcessor \underline{\textbf{\textit{rm d1+d2}}}} & {\fontsize{7.6}{8}\selectfont UniProcessor \underline{\textbf{\textit{rm d2+d1}}}} \\
Degradation & {\fontsize{6}{8}\selectfont PSNR / SSIM / LPIPS} & {\fontsize{6}{8}\selectfont PSNR / SSIM / LPIPS} & {\fontsize{6}{8}\selectfont PSNR / SSIM / LPIPS} & {\fontsize{6}{8}\selectfont PSNR / SSIM / LPIPS} & {\fontsize{6}{8}\selectfont PSNR / SSIM / LPIPS} & {\fontsize{6}{8}\selectfont PSNR / SSIM / LPIPS} & {\fontsize{6}{8}\selectfont PSNR / SSIM / LPIPS} \\
\midrule[0.15em]
Low-light + resize	&	19.3	/	0.63	/	0.39	&	19.3	/	0.63	/	0.38	&	20.1	/	0.64	/	0.38 &	21.2	/	0.60	/	0.42	&	19.4	/	0.64	/	0.37	&	\color{red}{22.9}	/	\color{red}{0.68}	/	\color{red}{0.36}	&	\color{blue}{22.8	/	0.67	/	0.37}	\\
{\fontsize{8.4}{8}\selectfont Color saturate + resize}	&	18.4	/	0.63	/	0.51	&	18.4	/	0.63	/	0.50	&	19.1	/	0.63	/	0.47 &	20.6	/	0.58	/	0.47	&	18.5	/	0.63	/	0.49	&	\color{blue}{22.0	/	0.65	/	0.41}	&	\color{red}{22.5}	/	\color{red}{0.65}	/	\color{red}{0.40}	\\
Rain + low-light	&	21.8	/	0.83	/	0.13	&	22.2	/	0.85	/	0.11	&	23.3	/	0.86	/	0.10 &	23.0	/	0.87	/	0.09	&	19.3	/	0.58	/	0.38	&	\color{blue}{28.4	/	0.89	/	0.08}	&	\color{red}{28.4}	/	\color{red}{0.90}	/	\color{red}{0.07}	\\
{\fontsize{8.4}{8}\selectfont Color saturate + noise}	&	20.8	/	0.93	/	0.20	&	21.1	/	0.93	/	0.19	&	21.6	/	0.94	/	0.15 &	22.6	/	0.87	/	0.26	&	20.8	/	0.93	/	0.20	&	\color{blue}{24.4	/	0.93	/	0.14}	&	\color{red}{33.9}	/	\color{red}{0.98}	/	\color{red}{0.02}	\\

Over bright + noise	&	16.1	/	0.79	/	0.17	&	16.1	/	0.79	/	0.18	&	18.0	/	0.81	/	0.17 &	16.0	/	0.78	/	0.18	&	17.8	/	0.72	/	0.22	&	\color{red}{19.6}	/	\color{red}{0.83}	/	\color{red}{0.15}	&	\color{blue}{18.0	/	0.73	/	0.21}	\\
\bottomrule[0.15em]
\end{tabular}}
\end{center}
\vspace{-2.5em}
\end{table}

\paragraph{Quantitative results of multi-degradation processing.}
As shown in Table \ref{tab2}, the results of \underline{\textbf{\textit{rm d1}}} and \underline{\textbf{\textit{rm d2}}} have obvious distinctions, due to the remained degradations are different.
Moreover, the better results in \underline{\textbf{\textit{rm d1}}} and \underline{\textbf{\textit{rm d2}}} are better than Restormer \cite{zamir2022restormer} and PromptIR \cite{potlapalli2023promptir}, which manifests the effectiveness of UniProcessor.
Furthermore, after sequential processing (\underline{\textbf{\textit{rm d1+d2}}} and \underline{\textbf{\textit{rm d2+d1}}}), the performance can be greatly improved compared to the smaller improvement of PromptIR with twice inference.
The order affects the sequential process depending on tasks and the 1st stage results.
For that both the first stage processes are well, the order effect on the results is small. For cases that the first degradation process strongly destroys the feature of another distortion, as processing low-light 1st for low-light+noise, the order strongly affects the second results.

\begin{table*}[t]
    \centering
    \caption{Comparisons under All-in-one restoration setting \cite{li2022all,potlapalli2023promptir}: single model trained on a combined set of images originating from different degradation types.}
    \vspace{-0.3em}
    \setlength{\tabcolsep}{0.45em}
    \renewcommand{\arraystretch}{1.1}
    \scalebox{0.6}{
    \begin{tabular}{l|c c c c c x{0} | x{60}}
        \toprule
        \multirow{2}{*}{Method} & Dehazing & Deraining & \multicolumn{3}{c}{Denoising on BSD68 dataset \cite{martin2001database_bsd}} & & \multirow{2}{*}{Average} \\
        & on SOTS \cite{li2018benchmarking} & on Rain 100L \cite{fan2019general} & $\sigma=15$ & $\sigma=25$ & $\sigma=50$ & & \\
        \midrule
        BRDNet \cite{tian2020image} & 23.23/0.895 & 27.42/0.895 & 32.26/0.898 & 29.76/0.836 & 26.34/0.836 & & 27.80/0.843 \\
        LPNet \cite{gao2019dynamic} & 20.84/0.828 & 24.88/0.784 & 26.47/0.7782 & 24.77/0.748 & 21.26/0.552 & & 23.64/0.738 \\
        FDGAN \cite{dong2020fd} & 24.71/0.924 & 29.89/0.933 & 30.25/0.910 & 28.81/0.868 & 26.43/0.776 & & 28.02/0.883 \\
        MPRNet \cite{Zamir_2021_CVPR_mprnet} & 25.28/0.954 & 33.57/0.954 & 33.54/0.927 & 30.89/0.880 & 27.56/0.779 & & 30.17/0.899 \\
        DL \cite{fan2019general} & 26.92/0.391 & 32.62/0.931 & 33.05/0.914 & 30.41/0.861 & 26.90/0.740 & & 29.98/0.875 \\
        AirNet \cite{li2022all} & 27.94/0.962 & 34.90/0.967 & 33.92/0.933 & 31.26/0.888 & 28.00/0.797 & & 31.20/0.910 \\
        PromptIR \cite{potlapalli2023promptir} & \textcolor{blue}{30.58/0.974} & \textcolor{blue}{36.37/0.972} & \textcolor{blue}{33.98/0.933} & \textcolor{blue}{31.31/0.888} & \textcolor{blue}{28.06/0.799} & & \textcolor{blue}{32.06/0.913} \\
        UniProcessor (Ours) & \textcolor{red}{31.66/0.979} & \textcolor{red}{38.17/0.982} & \textcolor{red}{34.08/0.935} & \textcolor{red}{31.42/0.891} & \textcolor{red}{28.17/0.803} & & \textcolor{red}{32.70/0.918} \\

        \bottomrule
    \end{tabular}
    }
    \label{tab3:gen}
\end{table*}

\paragraph{Results on an all-in-one restoration benchmark.} We further conduct an experiment following the all-in-one settings proposed in AirNet \cite{li2022all} and PromptIR \cite{potlapalli2023promptir}.
As shown in Table \ref{tab3:gen}, our UniProcessor achieves better performance compared to other state-of-the-art models, which further demonstrates the superiority and generality of the proposed method.

\begin{table}[t]
    \centering
    \captionof{table}{Ablation study results of UniProcessor. We report the PSNR/SSIM/LPIPS results on two tasks including rain streak removal, and low-light enhancement.}
    \vspace{-0.2em}
    \label{tab:ablation}
    \begin{minipage}{0.48\textwidth}\centering
    \scalebox{0.7}{
    \begin{tabular}{l|cc}\toprule
    \textbf{Method} &\textbf{Rain streak} &\textbf{Low-light}\\\midrule
    only text encoder &27.76/0.873/0.092 & 23.85/0.836/0.152 \\
    w/o Q-Former &\underline{28.23}/\underline{0.884}/\underline{0.087} & \underline{24.05}/\underline{0.840}/\underline{0.143} \\
    UniProcessor &\textbf{28.67/0.890/0.082} & \textbf{24.23/0.846/0.138}\\
    \bottomrule
    \end{tabular}}
    \vspace{0.1em}
    \subcaption{Ablation study for the context control module}\label{tab:4a}
    \end{minipage}
    \hfill
    \begin{minipage}{0.48\textwidth}\centering
    \scalebox{0.6}{
    \begin{tabular}{l|cc}\toprule
    \textbf{Method} &\textbf{Rain streak} &\textbf{Low-light}\\\midrule
    w/o CIM &27.30/0.862/0.101 & 23.03/0.832/0.156 \\
    level 5 &28.30/0.877/0.097 & 23.65/0.839/0.150 \\
    level 5+4 &\underline{28.44}/\underline{0.882}/\underline{0.088} & \underline{24.10}/\underline{0.842}/\underline{0.141} \\
    level 5+4+3 &\textbf{28.67/0.890/0.082} & \textbf{24.23/0.846/0.138}\\
    \bottomrule
    \end{tabular}}
    \vspace{0.1em}
    \subcaption{Ablation study for the block position of the context interaction module.}\label{tab:4b}
    \end{minipage}
    \vspace{-2em}
\end{table}

\subsection{Ablation Study}
We further conduct ablation studies for the UniProcessor as shown in Table \ref{tab:4a} \& \ref{tab:4b}.
Table \ref{tab:4a} demonstrates the ablation results for the context control module.
It can be observed that when only using text encoder to extract control information, the performance decreases obviously, and without using Q-Former to encode subject-aligned image representation, the performance also reduces.
Table \ref{tab:4b} demonstrates the ablation results for the context interaction module.
We observe that without CIM, the performance decreases a lot.
Adding CIM to shallow level can improve the performance but also increase computational cost.
Thus, we only add the CIM to level-5, level-4 and level-3 stages.

\begin{table}[t]
    \centering
    \captionof{table}{Comparisons of the computational overhead of UniProcessor.}
    \label{tab:ablation2}
    \begin{minipage}{0.44\textwidth}\centering
    \scalebox{0.7}{
    \begin{tabular}{l|c c c}
        \toprule
        variant & GMACs & PSNR \\
        \midrule
        u. swin as sa & 13.50 & 39.77 \\
        r. ca w/ sa & 13.40 & 39.75 \\
        r. sa w/ ca & \textbf{11.94} & \underline{39.78} \\
        UniProcessor & \underline{12.67} & \textbf{39.81} \\
        \bottomrule
    \end{tabular}}
    \vspace{0.2em}
    \subcaption{Computational overhead ablation of the Processor backbone of the UniProcessor.}\label{tab:5a}
    \end{minipage}
    \hfill
    \begin{minipage}{0.52\textwidth}\centering
    \scalebox{0.65}{
    \begin{tabular}{l|ccc}\toprule
    Model & processor & processor+CCM & processor+CCM+VQA \\
    \midrule
    GMacs & \textbf{153.40} & \underline{239.83} & 678.52 \\
    \midrule
    \midrule
    Model & MPRNet \cite{Zamir_2021_CVPR_mprnet} & PromptIR \cite{potlapalli2023promptir} & UniProcessor (w/o VQA) \\
    \midrule
    GMacs & 761.00 & \textbf{158.40} & \underline{239.83} \\
    \bottomrule
    \end{tabular}}
    \vspace{0.5em}
    \subcaption{Computational overhead comparisons with other models.}\label{tab:5b}
    \end{minipage}
\vspace{-2em}
\end{table}

We further conduct ablation experiments for the computational overhead.
We first test the performance and the computing overhead of the proposed processor backbone on SIDD dataset \cite{sidd}, and show the results in Table \ref{tab:5a}.
``u. swin as sa'' represents using swin transformer as the spatial attention, \textit{i.e.}, replacing the ConvFormer block with the swin transformer block.
``r. ca w/ sa'' and ``r. sa w/ ca'' indicate replacing channel attention with spatial attention and replacing spatial attention with channel attention, respectively, \textit{i.e.}, repeating spatial attention and repeating channel attention twice for each block, respectively.
We can observe that, compared to CSformer \cite{duan2023masked}, replacing swin blocks with large-kernel convolutional blocks can effectively improve the performance.
Moreover, we observe that the channel attention (ca) and spatial attention (sa) (large-kernel convolution layer) together contribute to the final improvement, and replacing one module with another module will decrease the performance.

Furthermore, we compare the computational overhead for different modules and models.
UniProcessor contains three modules, which include a VQA module, a context control module (CCM), a processor backbone. 
As shown in Table \ref{tab:5b}, the GMacs for the processor, processor+CCM, processor+CCM+VQA are 153.4, 239.83, 678.52, respectively. 
The LLM in the VQA module is the main overhead.
Since we can only use processor+CCM for processing, thus the overall processing GMac for UniProcessor is 239.83, which is comparable to other models (GMacs for PromptIR \cite{potlapalli2023promptir}, MPRNet \cite{Zamir_2021_CVPR_mprnet} are 158.4, 761).

\vspace{-0.8em}
\section{Conclusion}
\vspace{-0.2em}
In this work, we present a text-induced unified image processor, termed UniProcessor, for all-in-one image processing.
UniProcessor first has the ability to perceive low-level degradations and perform quality or degradation-related VQA, which can be used for generating the low-level subject prompt for subsequent processing procedure.
Moreover, to achieve controllable and unified image processing, we develop a text-induced processor, which encodes degradation-specific information from input image and subject text prompt, and incorporates the manipulation prompt into the degradation-aware embedding to obtain context control information.
The control embedding is interacted with the processor backbone to achieve controllable and unified image processing.
Extensive experimental results demonstrate that UniProcessor can well process 30 degradations in one model which outperforms other competing methods, and achieve the ability to process individual distortion in an image with multiple degradations.

\clearpage  

%
%
\bibliographystyle{splncs04}
\bibliography{main}
\end{document}